\documentclass[conference]{IEEEtran}
\IEEEoverridecommandlockouts
\usepackage{cite}
\usepackage{amsmath,amssymb,amsfonts}
\usepackage{algorithmic}
\usepackage{graphicx}
\usepackage{textcomp}
\usepackage[linesnumbered,ruled,vlined]{algorithm2e}
\usepackage{amssymb}
\usepackage{amsmath}
\usepackage{lmodern}
\usepackage{siunitx}
\usepackage{booktabs}
\usepackage{etoolbox}
\usepackage{array}
\newcolumntype{?}{!{\vrule width 1pt}}
\usepackage{mathtools}
\usepackage{booktabs}
\usepackage{siunitx}
\usepackage{multirow}
\usepackage{footmisc}
\usepackage{hyperref}

\usepackage{xcolor}
\def\BibTeX{{\rm B\kern-.05em{\sc i\kern-.025em b}\kern-.08em
    T\kern-.1667em\lower.7ex\hbox{E}\kern-.125emX}}
\begin{document}

\title{On Brightness Agnostic Adversarial Examples Against Face Recognition Systems}

\author{\IEEEauthorblockN{Inderjeet Singh}
\IEEEauthorblockA{\textit{NEC Corporation} \\
inderjeet78@nec.com}
\and
\IEEEauthorblockN{Satoru Momiyama}
\IEEEauthorblockA{\textit{NEC Corporation} \\
satoru-momiyama@nec.com}
\and
\IEEEauthorblockN{Kazuya Kakizaki}
\IEEEauthorblockA{\textit{NEC Corporation, University of Tsukuba} \\
kazuya1210@nec.com}
\and
\IEEEauthorblockN{Toshinori Araki}
\IEEEauthorblockA{\textit{NEC Corporation} \\
toshinori\_araki@nec.com}
}

\maketitle

\begin{abstract}
This paper introduces a novel adversarial example generation method against face recognition systems (FRSs). An adversarial example (AX) is an image with deliberately crafted noise to cause incorrect predictions by a target system. The AXs generated from our method remain robust under real-world brightness changes. Our method performs non-linear brightness transformations while leveraging the concept of curriculum learning during the attack generation procedure. We demonstrate that our method outperforms conventional techniques from comprehensive experimental investigations in the digital and physical world. Furthermore, this method enables practical risk assessment of FRSs against brightness agnostic AXs. 
\end{abstract}


\section{Introduction}
The recent advancement in Adversarial Machine Learning (AML) has discovered that state-of-the-art (SOTA) Deep Learning (DL) models are vulnerable to well-designed input samples called \textit{Adversarial Examples} (AXs) \cite{goodfellow2014explaining}. The vulnerability to AXs becomes a significant risk for applying deep neural networks in safety-critical applications like Face Recognition Systems (FRSs). Face Recognition is a process of validating a claimed identity based on the image of a face. An adversary can conveniently attack practical FRSs from the digital and the physical world, e.g., in ID photo-matching systems \cite{folego2016cross}. 

In digital attacks, the digital adversarial noise is directly added to the target digital image. In the physical attacks, digital AX is transferred to the physical world (by printing, etc.) and then used to attack a target system. The generated AXs can be white-box, gray-box, or black-box depending on whether they are generated leveraging complete, partial, or no access, respectively, to the target system's information. Various digital and physical perturbations affect these AXs because the AXs are typical images with a few highly correlated adversarial features with the target ML model's predictions. The perturbations can be in color corrections, contrast change, hue shift, and brightness changes. The brightness change is one of the critical parameters, causing a significant change in AX's performance.

\begin{figure*}[t]
\centering
    \includegraphics[width=\textwidth]{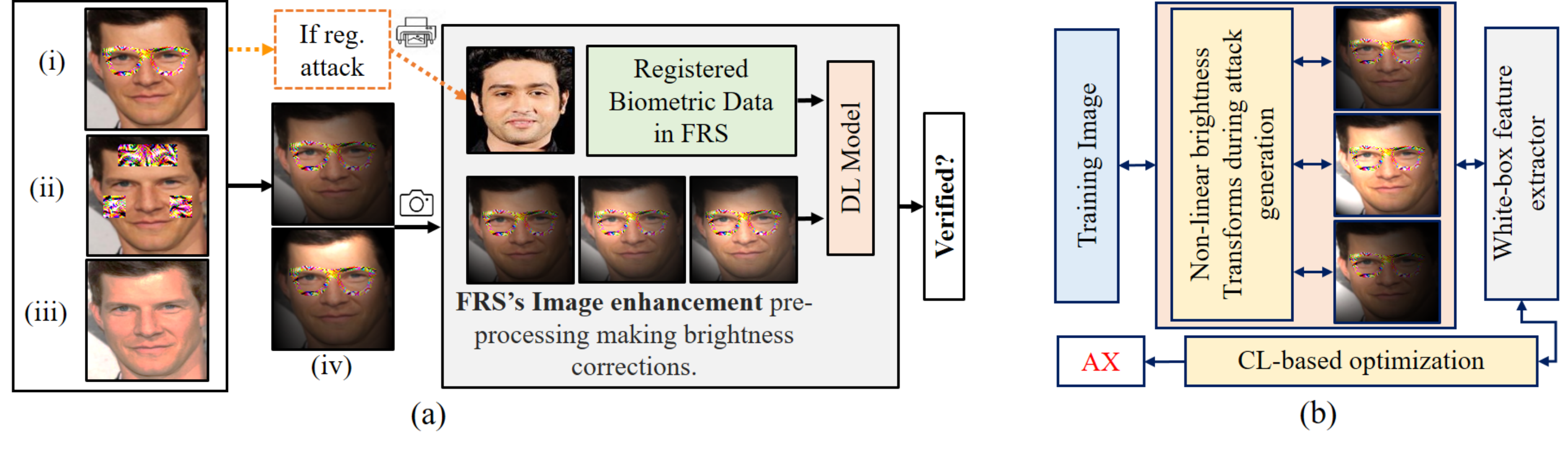}
    \caption{\label{fig:flowchart} (a) illustrates an example of practical FRS and brightness corrections in input face images \cite{vggface2} during FRS's pre-processing. (i), (ii), and (iii) illustrate eyeglass, sticker, and imperceptible noise AXs. (iv) represents face images taken under variable brightness. (b) demonstrates an outline of our proposed method.}
\end{figure*}

The practical risk assessment of the FRSs scans the possible vulnerabilities of the ML model used in the FRS from different kinds of AXs. However, the brightness changes weaken the AX, making it non-suitable for the practical risk assessment of the target system. Thus powerful AXs robust to the brightness changes must be adopted. When an AX succeeds even in altering brightness environments, it is called \textit{brightness agnostic AX}. In practical scenarios, brightness changes \textit{non-linearly}. In the digital world, \textit{non-linear} brightness changes occur due to the use of image enhancement techniques \cite{ying2017new}\cite{ren2015novel} by FRSs for improved performance, which can be seen in \texttt{Fig.\ref{fig:flowchart}a}. In the physical world, four primary factors cause brightness changes: printer specification, printing surface properties, environmental illumination conditions, and camera specifications. 

Yang et al. \cite{yang} proposed an adversarial example generation method based on random transformations of image brightness. 
They reduced the overfitting, thereby improving black-box transferability of generated attacks, by applying linear brightness transformations on the training\footnote{\label{training}In the context of AML, the trainable parameters are the adversarial noise pixels in the input image, optimized using an attack generation method. Thus we call the input image being optimized for adversarial objective as training image in the present setting.} image optimized for an adversarial objective. However, \cite{yang} did not evaluate the robustness of the generated attacks in changing brightness conditions. Also, they assumed only linear brightness changes. Additionally, the FRSs were not considered in their evaluation. Therefore, in this work, in addition to our proposed method, we also evaluate (1) the robustness of the attacks generated from their method in the changing brightness conditions, (2) the improvement in the black-box transferability, and (3) performance for FRSs.

\textbf{Our main contributions are:} We propose a \textit{novel Curriculum Learning (CL)-based method} for generating AXs robust to \textit{real-world brightness changes}. To our best knowledge, this is the very first attempt for generating brightness agnostic adversarial attacks. We conduct extensive experiments on four SOTA face verification models under a well-known PGD (Projected Gradient Descent) attack \cite{madry} setting. We evaluated the \textit{white-box} and \textit{black-box} attack performance in the \textit{digital} as well as the \textit{physical world}. We also evaluate our method against the FRSs deployed with adversarial defenses in the pre-processing pipeline.

\section{Our method for generating brightness agnostic AXs}
The proposed method \texttt{[Alg.\ref{alg:1}]} yields \textit{non-linear} brightness changes during the attack generation process, as it can be seen in \texttt{Fig.\ref{fig:flowchart}b}. The non-linear change in the brightness during attack generation makes the generated AXs robust to them during inference. To better optimize the challenging non-linear brightness changes, our method uses the concept of \textit{\textbf{CL}} for generating \textit{\textbf{B}rightness \textbf{A}gnostic AXs} in the \textbf{PGD} attack setting; thus, we call our method a CL-BA-PGD attack. CL is an approach proposed by \cite{El93} in which training difficulty is gradually increased while training DL models for better performance.


\begin{algorithm*}[t]
\SetAlgoLined
\caption{CL-BA-PGD Algorithm for Adversarial Patch Attacks}
 \label{alg:1}
\KwIn{Source image $X^s$ of identity $s$; target image $X^t$ of identity $t$; face-matcher $f$; adversarial loss function $J_{adv}$; random noise $\delta$; patch mask $M_p$; brightness mask $M_b$; stopping criteria $T$; step functions $g_1$ \& $g_2$; batch constant $N$; similarity constant $K$; number of brightness ensembles $N_b$; learning rate $\alpha$.}
\KwOut{Brightness agnostic patch adversarial example $X^{adv} = X_{T-1}^{adv}$}

$X_{0}^{adv} \leftarrow X^s\cdot M_p^{'} + \delta \cdot M_p$,
$l_{0} \leftarrow 1$, $h_{0} \leftarrow 1$, $p \leftarrow 0$, $loss^{cum}_0\leftarrow0$ \\
 \For{i=0 to T-1 \do}{
  \For{j=0 to $N_{b}-1$ \do}{ $X_{i,j} \longleftarrow CNBT_{j}\left(X_{i}^{adv};p;M_p;M_{p}^{'};M_b;M_{b}^{'};l_i; h_i\right) =  \left(Y_j \cdot \left(BT\left(X_i^{adv}\cdot M_p\right)+ X_{i}^{adv}\cdot M_{p}^{'}\right)\right)\cdot \left(M_{b,j} \cdot X_u + M_{b,j}^{'}\right)$
  }
  $X_{i+1}^{adv} \longleftarrow clip_{0-1}\left(X_i^{adv} - \alpha \cdot sign\left(\sum_{j=0}^{N_b-1}\nabla J_{adv}(f(X_{i,j}), f(X^t))\right)\right)$ \\ 
  $loss^{cum}_{i+1}= loss^{cum}_{i} + \frac{\sum_{j=0}^{N_b-1} J_{adv}(f(X_{i,j}), f(X^t))}{N_b}$\\
  $l_{i+1} \longleftarrow g_{1}(l_{i})$; $h_{i+1} \longleftarrow g_{2}(h_{i})$  \\ 
  \If{$i\neq0$ and $i \% N = 0$}{
  $p = max\left(0, \left(K- \frac{loss^{cum}_i}{N} \right)\right)$\\
  $loss^{cum}_{i+1}= \frac{\sum_{j=0}^{N_b-1} J_{adv}(f(X_{i,j}), f(X^t))}{N_b}$
  } 
 }

\end{algorithm*}

To generate attacks using our algorithm \texttt{[Alg.1]}, non-linear brightness transformations $CNBT_{j}()$ are applied to the training\footref{training} image $X_{i}^{adv}$ after the initialization \texttt{[Alg.\ref{alg:1}; 1]}. The transformations \texttt{[Alg.\ref{alg:1}; 4]} are applied while regulating the optimization difficulty based on the loss $J_{adv}$. The loss $J_{adv}$ in gradient descent setting calculates the inverse of cosine similarity between the predictions of $f$ for $X^s$ and $X^t$ for the impersonation attacks and simply similarity for dodging attacks. For impersonation attacks, the adversary with identity $s$, tries to mimic the deep features of target identity $t$. For dodging attacks, $s$ and $t$ are same because adversary tries to minimize the similarity from its clean image's deep features. The predefined step-functions $g_1$ and $g_2$ change lower $l_i$ and upper $h_i$ limits for a uniform random variable $X_u \sim U(l_i,h_i)$, thus controlling the non-linear brightness changes.

The function $BT$ changes the brightness of an image tensor $X$ as $BT(X)= X_u \cdot X$ with probability $p$. The 0-1 mask $M_{b,j}$ with the same dimensions as the $X_{i}^{adv}$, randomly chooses a rectangular area $\mathcal{R}_b$ inside $X_{i}^{adv}$ in each $j^{th}$ iteration to scale brightness by $X_u$. Thus, $(M_{b,j})_{(m,n)}=1$ if $(m,n) \in \mathcal{R}_b$ and $(M_{b,j})_{(m,n)}=0$ if $(m,n) \notin \mathcal{R}_b$. The patch masks $M_p$ is used to separate the predefined patch area inside $X_{i}^{adv}$. Also, $M_{p}^{'} = I_1-M_p$ and $M_{b,j}^{'} = I_1-M_{b,j}$ where $I_1$ is all one matrix. The random variable $Y_j \sim N(\mu_{j}, \sigma_{j})$ follows Gaussian distribution. 

The PGD updates are then performed on $X_{i}^{adv}$ following \texttt{[Alg.\ref{alg:1}; 6]}. Note that it is assumed that images are normalized in the [0,1] range. The parameters responsible for the CL are updated in the subsequent steps \texttt{[Alg.\ref{alg:1}; 7,8,9,11]} following the idea of gradually changing the amount of brightness changes depending upon $J_{adv}$. The parameter $K$ is loss function specific and serves to provide a margin for the minimum values of the $p$ parameter.

\subsection{Sorting optimization difficulty}
We define the optimization difficulty in the $i^{th}$ iteration of the attack generation process as directly proportional to $\Delta \mathcal{L}_{V_{B}^{\cdot}}^{k}$, which is the change in the adversarial loss caused by variation in the brightness $V_{B}^{\cdot}$ of input image due to application of a ($\cdot$)-type transformation. The large change in the adversarial loss causes significant variations in gradient-based methods' descent direction, making the optimization process challenging. Also, $\Delta \mathcal{L}_{V_{B}^{\cdot}}^{k}$ is calculated for $k$ training\footref{training} images for a DL model $f$ trained for $t \leq T$ iterations. 

\begin{figure*}[h!]
\centering
    \includegraphics[width=\textwidth]{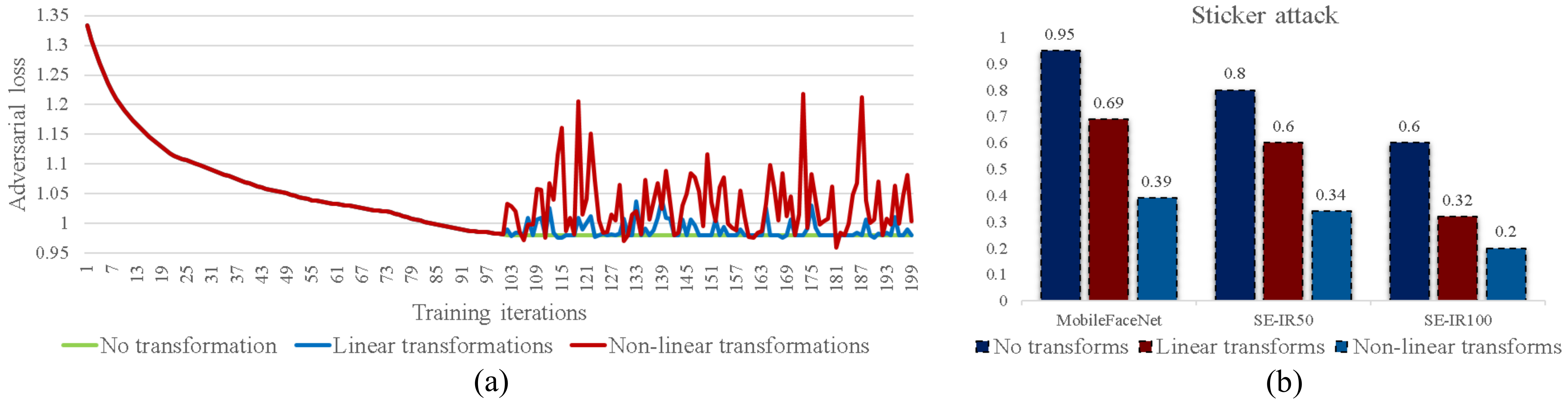}
    \caption{\label{fig:advloss1}(a) depicts the variation of the adversarial loss for the trained (till 100 iterations) MobileFaceNet \cite{chen2018mobilefacenets} when subjected to no, linear, and non-linear brightness transformations. (b) demonstrates the performance (ASR) reduction due to the linear and non-linear brightness transformations for sticker attacks.}
\end{figure*}

We applied linear and non-linear brightness transformations to a face image with adversarial eyeglass patch noise to assess the increased training\footref{training} difficulty. We saw from \texttt{Fig.\ref{fig:advloss1}a} that maximum variations in the adversarial loss (hence the optimization difficulty) was caused by non-linear brightness transformations followed by linear and no brightness transformations, i.e. $\Delta \mathcal{L}_{V_{B}^{nl}}^{k} > \Delta \mathcal{L}_{V_{B}^{l}}^{k} >\Delta \mathcal{L}_{V_{B}^{0}}^{k} = 0$. Thus our hypothesis is that increased optimization difficulty reduces the chances of convergence to optimal solutions thereby reducing attack success probability.



To investigate the effect of the brightness changes and adversarial loss variations on the adversarial potential of successful AXs, we evaluated the reduction in attack success rate (ASR) due to linear \cite{yang} and non-linear brightness transformations. ASR is the fraction of AXs which successfully fooled the DL model during inference. After evaluating eyeglass, sticker, and imperceptible noise attacks, the results for the ASR confirmed our hypothesis that non-linear brightness transformations cause a significant reduction in the ASR compared to the linear transformations.

\section{Experimental Setup}\label{exp_setup}
For an adequate assessment, we considered \textit{four} SOTA feature extractors: Residual Network (ResNet50)  \cite{he2016deep}, MobileFaceNet \cite{chen2018mobilefacenets}, and Squeeze-and-Excitation Inception Residual Networks (SE-IR50, SE-IR100) \cite{hu2018squeeze}; trained on the VggFace2 data \cite{vggface2} using the arcface loss. The test accuracy on Vggface2 data of trained ResNet50, MobileFaceNet, SE-IR50, and SE-IR100 was found as $99.03\%$, $99.17\%$, $99.01\%$, and $99.02\%$, respectively. For each feature extractor, we implemented a simple PGD attack \cite{madry}, the existing method \cite{yang}, our method without CL, and our method with CL \texttt{[Alg.\ref{alg:1}]}.

We implement our algorithm for the patch and the imperceptible noise attacks while evaluating in digital and physical worlds. The patch attacks were generated using \texttt{[Alg.1]}. For the imperceptible noise attacks, the adversarial noise $\delta$ with size constraints ($\delta \leq (4/255)^{th}$ of input image's pixel range) was distributed over the entire area of the face image. In this case, we implemented \texttt{[Alg.1]} by changing \texttt{[Alg.\ref{alg:1}; 4]} with $CNBT_{j}\left(\cdot\right) =  Y_j \cdot \left(BT\left(X_i^{adv} \cdot M_{b,j}\right){}+ X_{i}^{adv}\cdot M_{b,j}^{'}\right)$.

To evaluate the white-box and black-box performance of the generated attacks, the white-box attacks were generated and tested directly on the target FRS \texttt{(Tab.1)}. In contrast, the black-box attacks were generated using a surrogate FRS and tested on the target FRS \texttt{(Tab.2)} to evaluate the transferability of generated attacks.

For adversarial patch attacks, we considered an eyeglass patch (\texttt{Fig.\ref{fig:flowchart}a(i)}) and a sticker patch (\texttt{Fig.\ref{fig:flowchart}a(ii)}). For the practical evaluation of the generated attacks (100 AXs for each case) in the \textit{digital domain}, the mean ASR for each AX was calculated after applying \texttt{[Alg.\ref{alg:1}; 4]} transformations 100 times to simulate the practical brightness variations. For the evaluation in the \textit{physical world}, the following steps were followed: (1) Generate digital AX.(2) Transfer it to the physical world by printing at 9 different brightness levels. (3) Capture the printed AXs from various angles. (4) Clean the captured data. We got approximately 20 images for each captured image. (5) Feed the data to the MTCNN face detection and alignment \cite{xiang2017joint}. (6) Feed the preprocessed data to the target feature extractor and check the predictions.

\begin{table*}[t]
    \centering
    \sisetup{detect-weight,mode=text}
    \renewrobustcmd{\bfseries}{\fontseries{b}\selectfont}
    \renewrobustcmd{\boldmath}{}
    \newrobustcmd{\B}{\bfseries}
    \addtolength{\tabcolsep}{7.82pt}
     \begin{tabular}{c|c?c|c|c|c?c|c|c|c}
     \toprule
     Attack     &  White-box   & \multicolumn{4}{c?}{ Mean IASR } & \multicolumn{4}{c}{Mean DASR} \\ \cline{3-10}
     Type       &  model       & A1                & A2                & A3                & A4     & A1 & A2 & A3 &    A4     \\
     \midrule
                & ResNet50       & $0.39$   & $0.42 $   & $0.56 $   & \B 0.58   & $0.63 $ & $0.66 $ & $0.74 $ & \B 0.78    \\
     \cline{2-10}
     Eyeglass          & MobileFaceNet       & $0.44 $   & $0.51 $   & $0.53 $   & \B 0.55   &  $0.52 $ & $0.60 $ & $0.63 $ & \B 0.69    \\
     \cline{2-10}
     Attack            & SE-IR50       &   $0.41 $ & $0.48 $   & $0.51 $   & \B 0.60   & $0.56 $ & $0.65 $ & $0.62 $ & \B 0.73     \\
     \cline{2-10}
                & SE-IR100       & $0.29 $   & $0.37 $   & $0.40 $   & \B 0.45   &   $0.42 $ & $0.49 $ & $0.48 $ & \B 0.53     \\
     \midrule
                & ResNet50       &$0.54 $    & $0.69 $   & $0.72 $   & \B 0.86   & $0.52 $ & $0.69 $ & $0.85 $ & \B 0.95    \\
     \cline{2-10}
     Sticker          & MobileFaceNet       &  $0.54 $  & $0.66 $   & $0.68 $   & \B 0.75  &  $0.48 $ & $0.62 $ & \B 0.68   & $0.66 $ \\
     \cline{2-10}
     Attack            & SE-IR50       &   $0.49 $ & $0.59 $   & $0.64 $   & \B 0.69   &  $0.42 $ & $0.48 $ & $0.52 $ & \B 0.60   \\
     \cline{2-10}
                & SE-IR100       &  $0.43 $  & $0.55 $   & \B 0.65    & $0.60 $  &  $0.41 $ & $0.50 $ & $0.63 $ & \B 0.70   \\
     \midrule
                & ResNet50       &$0.34 $    & $0.43 $   & \B 0.51    & $0.50 $  & $0.44 $ & $0.55 $ & $0.44 $ & \B 0.62     \\
     \cline{2-10}
     Imperceptible          & MobileFaceNet       &  $0.30 $  & $0.33 $   & $0.36 $   & \B 0.45   &  $0.38 $ & $0.44 $ & $0.52 $ & \B 0.63  \\
     \cline{2-10}
     Noise Attack            & SE-IR50       &   $0.21 $ & $0.25 $   & $0.38 $   & \B 0.52   &  $0.47 $ & $0.57 $ & $0.60 $ & \B 0.63  \\
     \cline{2-10}
                & SE-IR100       & $0.24 $   & $0.26 $   & $0.28 $   & \B 0.34   &   $0.16 $ & $0.23 $ & $0.32 $ & \B 0.52  \\
     \bottomrule
     \end{tabular}
     \vspace{0.1cm}
     \caption{The mean ASR of the \textit{\textbf{white-box}} attacks in the digital domain. DASR \& IASR are Dodging \& Impersonation ASRs. $A_1$, $A_2$, $A_3$, and $A_4$ stands for naive, existing, our method w/o CL, and our CL-based methods, respectively.}
      \label{tab:wb_imp_dig}
   \end{table*}

\section{Results}   
\texttt{Tab.\ref{tab:wb_imp_dig}} and \texttt{Tab.\ref{tab:bb}} shows the results for white-box and black-box ASRs, respectively, in the digital domain. The mean ASR is calculated for 100 AXs after applying the transformations mentioned in section \ref{exp_setup} on each AX. Our method with CL results in a significantly higher ASR as compared to the existing method \cite{yang} and the naive method. Also, the existing method achieves better results than the naive method. The effect of better optimization due to CL can also be seen from the increased ASR from \texttt{A3} to \texttt{A4} columns of \texttt{Tab.\ref{tab:wb_imp_dig}}. Our method also achieves better ASR for the digital black-box attacks \texttt{(Tab.\ref{tab:bb})}. However, in this case, the performance difference was not as significant as in the white-box setting. Additionally, sticker attacks were found to be having the highest ASRs (\texttt{Tab.\ref{tab:wb_imp_dig}} and \texttt{Tab.\ref{tab:bb}}) due to the larger area for the adversarial noise region and absence of imperceptible size constraints.

\begin{table*}[t]
    \centering
    \sisetup{detect-weight,mode=text}
    \renewrobustcmd{\bfseries}{\fontseries{b}\selectfont}
    \renewrobustcmd{\boldmath}{}
    \newrobustcmd{\B}{\bfseries}
    \addtolength{\tabcolsep}{5.50pt}
     \begin{tabular}{c|c|c?c|c|c|c?c|c|c|c}
      \toprule
      Attack&Surrogate& Black-box& \multicolumn{4}{c?}{ Mean IASR }& \multicolumn{4}{c}{Mean DASR} \\ \cline{4-11}
      Type&  model &    models   & A1 & A2 & A3 & A4 & A1 & A2 & A3 & A4 \\
     \midrule
                    & $M_1$ & $M_2,M_3$     & 0.07  & \B0.08& 0.05  & \B 0.08   & 0.1   & 0.11  & 0.11  & \B 0.13   \\
     \cline{2-11}
     Eyeglass       & $M_2$ &$M_3,M_4$      & 0.13  & \B0.14& \B0.14& \B0.14    &0.16   & 0.20  & 0.20  &\B 0.23    \\
     \cline{2-11}
     Attack         & $M_3$ &$M_2,M_4$      & 0.16  & 0.16  & 0.15  & \B0.17    & 0.34  & 0.40  & 0.34  & \B0.45    \\
     \cline{2-11}
                    &$M_4$ &$M_2,M_3$       & 0.14  & 0.19  & 0.18  & \B0.21    & 0.25  & 0.35  & \B0.40& 0.37      \\
     \midrule
                    & $M_1$ &$M_2,M_3$      & 0.04  & 0.05  & \B0.06& 0.04      & 0.18  & 0.17  & 0.19  & \B0.20    \\
     \cline{2-11}
     Sticker        & $M_2$ &$M_3,M_4$      &  0.11 & 0.12  & 0.12  & \B0.13    & 0.24  & \B0.40& 0.27  & 0.35      \\
     \cline{2-11}
     Attack         & $M_3$ &$M_2,M_4$      & \B0.15& 0.14  & \B0.15& 0.13      &  0.38 & 0.40  & \B0.53& 0.50      \\
     \cline{2-11}
                    & $M_4$ &$M_2,M_3$      & 0.19  & 0.24& 0.21  & \B 0.25    & \B0.58& 0.46  & 0.52  & \B0.58    \\
     \midrule
                    & $M_1$ & $M_2,M_3$     &  0.08 & 0.10  & 0.11  & \B0.17    & 0.09  & 0.12  & 0.11  & \B0.18    \\
     \cline{2-11}
     Imperceptible  & $M_2$ &$M_3,M_4$      &  0.22 & 0.29  & 0.23  & \B0.32    &  0.19 & 0.18  & \B0.26& 0.18      \\
     \cline{2-11}
     Noise Attack   & $M_3$ &$M_2,M_4$      & 0.19  & 0.20  & 0.22  & \B0.34    &  0.34 & 0.36  & 0.32  & \B0.38    \\
     \cline{2-11}
                    & $M_4$ &$M_2,M_3$      &  0.15 & 0.14  & 0.27  & \B0.24    &  0.25 & 0.39  & 0.37  & \B0.43    \\
     \bottomrule

     \end{tabular}
     \vspace{0.1cm}
     \caption{ The ASR for the \textit{\textbf{black-box}} attacks in the digital domain. DASR \& IASR are Dodging \& Impersonation ASRs. $M_1$, $M_2$, $M_3$, and $M_4$ represents ResNet50, MobileFaceNet, SE-IR-50, and SE-IR-100 face feature extractors. $A_1$, $A_2$, $A_3$, and $A_4$ stands for naive, existing, our method w/o CL, and our CL-based methods, respectively.}
    \label{tab:bb}
   \end{table*}

The evaluation of the generated attacks in the \textit{physical domain} also exhibited similar patterns as \textit{digital white-box ASR} \texttt{(Tab.1)}. Our method achieves $ 24.67 \%$ and  $ 39.96 \%$ better mean ASR than the existing \cite{yang} and the naive PGD attack methods, respectively, for the eyeglass patch attack. Additionally, we evaluate the robustness of the brightness agnostic AXs against the model with JPEG compression \cite{jpeg_comp}, bit squeezing, and median blur defenses \cite{f_squeezing} in the pre-processing pipeline. These defenses do not directly cause brightness changes in the input images. After evaluation, we did not find sufficient evidence to validate the better ASR of the brightness agnostic AXs generated using our method against them.

\section{Conclusions}
This paper contributed a novel CL-based method for generating AXs robust to the practical brightness changes. While considering attacks from digital and physical worlds, we found that our approach significantly exceeds the conventional techniques in white-box and black-box settings from our detailed analysis of the dodging and impersonation attacks.
However, we did not find sufficient evidence for the superiority of our method against adversarial defenses that do not cause a direct change in the brightness of input images. A possible weakness of our approach is that it requires careful manual initialization of a few hyper-parameters responsible for CL that can directly affect attack performance. The generated attacks by our method enable practical risk assessment of the FRSs against such attacks. In the future, we would like to consider utilizing color space transformations, and assessing provided robustness improvements through adversarial training by our method.   

\bibliographystyle{IEEEtran}
\bibliography{main}

\end{document}